%% file: acl2023.tex
\pdfoutput=1

\documentclass[11pt,twocolumn]{article}

\usepackage[]{ACL2023}

\usepackage{times}
\usepackage{latexsym}

\usepackage[T1]{fontenc}

\usepackage[utf8]{inputenc}

\usepackage{microtype}
\usepackage{booktabs} 
\usepackage{siunitx}  

\usepackage{inconsolata}
\usepackage{graphicx}
\usepackage{xspace}
\usepackage{hyperref}
\usepackage{cleveref}
\usepackage{enumitem}
\usepackage[normalem]{ulem}
\crefformat{section}{\S#2#1#3}
\crefformat{subsection}{\S#2#1#3}
\crefformat{subsubsection}{\S#2#1#3}
\crefrangeformat{section}{\S#3#1#4 to~\S#5#2#6}
\crefmultiformat{section}{\S#2#1#3}{ and~\S#2#1#3}{, #2#1#3}{ and~#2#1#3}
\Crefformat{figure}{#2Fig.~#1#3}
\Crefmultiformat{figure}{Figs.~#2#1#3}{ and~#2#1#3}{, #2#1#3}{ and~#2#1#3}
\Crefformat{table}{#2Tab.~#1#3}
\Crefmultiformat{table}{Tabs.~#2#1#3}{ and~#2#1#3}{, #2#1#3}{ and~#2#1#3}
\Crefformat{appendix}{#2Appx.~\S#1#3}
\crefformat{algorithm}{Alg.~#2#1#3}
\Crefformat{equation}{#2Eq.~#1#3}
\newcommand{\modelname}{\textsc{EureQA}\xspace}
\newcommand{\fullname}{\texttt{\underline{E}}xtending \texttt{\underline{U}}nderlying \texttt{\underline{re}}asoning Chains in \texttt{\underline{QA}}\xspace}
\newcommand{\stitle}[1]{\vspace{1ex} \noindent{\bf #1}}

\title{Deceptive Semantic Shortcuts on Reasoning Chains: \\ How Far Can Models Go without Hallucination?}

\author{Bangzheng Li$^{1}$ \quad Ben Zhou$^{2}$ \quad Fei Wang$^{3}$ \quad Xingyu Fu$^{2}$ \quad Dan Roth$^{2}$ \quad Muhao Chen$^{1}$ \\
    $^{1}$University of California, Davis \quad
  $^{2}$University of Pennsylvania \\
  $^{3}$University of Southern California \\
  \texttt{bzhli@ucdavis.edu} 
}

\usepackage{graphicx}
\usepackage{etoolbox}
\newcommand{\insertfig}{
\vspace{-2em}
\includegraphics[width=\linewidth]{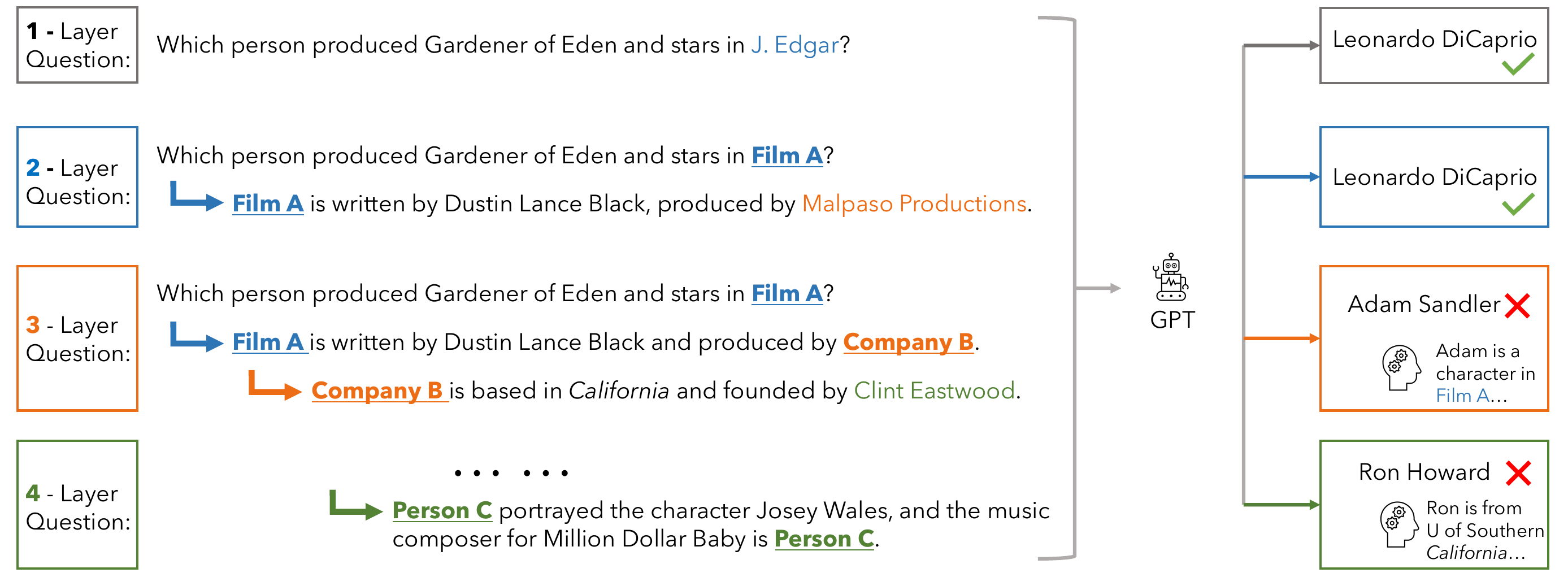}
\vspace{-1em}
\captionof{figure}{LLMs make errors when correct surface-level semantic cues -- entities -- are recursively replaced with descriptions, and the errors are likely related to \textit{token similarity}. GPT-3.5-turbo is used for this example.}
\label{fig:sample_question}
\vspace{2em}
}
\makeatletter
\apptocmd{\@maketitle}{\centering\insertfig}{}{}
\makeatother

\begin{document}
\maketitle

\input{content/0_abstract}
\input{content/1_introduction}

\input{content/3_dataset}

\input{content/4_expSetup}

\input{content/5_results}

\input{content/2_relatedWork}

\input{content/6_conclusion}
\bibliography{anthology,custom}
\bibliographystyle{acl_natbib}

\input{content/7_appendix}



\end{document}

%% file: content/0_abstract.tex
\begin{abstract}
Despite the high performances of large language models (LLMs) across numerous benchmarks, recent research has unveiled their suffering from hallucinations and unfaithful reasoning. This work studies a type of hallucination induced by semantic associations. We investigate to what extent LLMs take shortcuts from certain keyword/entity biases in the prompt instead of following correct reasoning paths. To quantify this phenomenon, we propose a novel probing method and benchmark called \modelname. \modelname{} is an entity-searching task where a model finds a missing entity based on described multi-hop relations with other entities. These deliberately designed multi-hop relations create deceptive semantic associations, and models must stick to the correct reasoning path instead of incorrect shortcuts to find the correct answer.
Experiments show that existing LLMs cannot follow correct reasoning paths and resist the attempt of greedy shortcuts, with GPT-4 only achieving 62\% accuracy. Analyses provide further evidence that LLMs rely on semantic biases to solve the task 
instead of proper reasoning, questioning the validity and generalizability of current LLMs' high performances.\footnote{Code and data are available at \url{https://github.com/VincentLeebang/eureqa}}

\end{abstract}

%% file: content/1_introduction.tex
\section{Introduction}

\begin{figure*}[t!]
\begin{center}
    \includegraphics[width=1\textwidth]{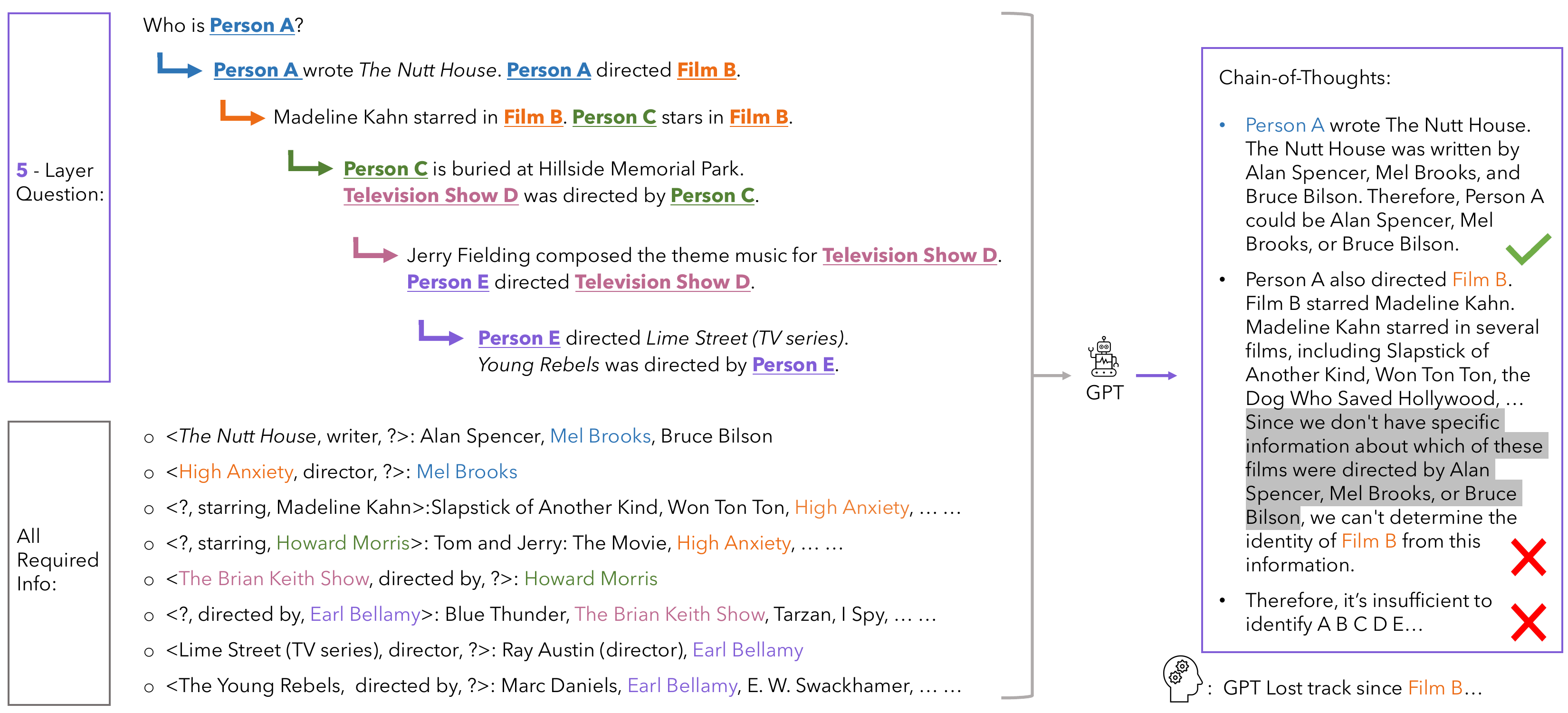}
    \caption{Even given all the required information needed for the question (selected information shown in the figure), GPT-4 still makes mistakes starting early layers (highlighted in grey). We only show partial output here. Notice that we give GPT few-shot prompts.}
    \label{fig:fig2}
    \vspace{-1em}
\end{center}
\end{figure*}

Recent progress of large language models (LLMs) has showcased their emergent abilities in a wide range of reasoning tasks, encompassing competence in program synthesis \cite{kuznia-etal-2022-less}, mathematical reasoning \cite{mishra-etal-2022-lila}, symbolic logic reasoning \cite{gaur-saunshi-2023-reasoning}, and common-sense reasoning \cite{feng-etal-2023-generic}. Innovations in inference methods have further improved language models' reasoning capabilities, either by generating intermediate steps toward the ultimate solution \cite{wei2022chain} or by decomposing complex inquiries into manageable sub-problems \cite{Zhou2022LearningTD, yao2023tree}. Research has further demonstrated that when incorporated with these techniques, LLMs have achieved exceptional results on multi-hop and complex QA benchmarks \cite{geva2021did}.

Nonetheless, it remains under-investigated whether the effectiveness of LLMs' reasoning pattern \cite{openai2023gpt4,geminiteam2023gemini} is genuinely based on a sensible reasoning path or if these models predominantly depend on semantic associations (i.e., word distributions from the pre-training data) to generate a plausible answer. \Cref{fig:sample_question} shows how ChatGPT\footnote{In this paper, we consistently utilize the gpt-3.5-turbo-0301 checkpoint for ChatGPT.} fails to find the missing entity as we gradually remove relevant semantic cues as the reasoning depth grows.
The original question (labeled as ``1 Layer'' in \Cref{fig:sample_question}) asks models to find an individual that satisfies two distinct conditions. It can be resolved by finding the set of people who are producers of the film ``Gardener of Eden'' and the set of cast members of ``J. Edgar'', and checking for the overlapping entity.
However, language models such as ChatGPT may not necessarily perform this sensible entity-seeking process. This is because the correct answer can simply (but incorrectly) be inferred by the high occurrence of ``Leonardo DiCaprio'' in the model's training data with ``Gardener of Eden'' and ``J. Edgar''.\footnote{In fact, if we prompt ChatGPT with only three keywords ``Gardner of Eden,'' ``J. Edgar'' and ``actor'' and let it generate anything, the outcome is almost always ``Leonardo DiCaprio.''} In such a scenario, ChatGPT may generate the correct response through this shortcut of semantic association. 
Using semantic shortcuts such as co-occurring entities works well on common test cases. Still, they limit systems' generalizability and robustness on cases not aligning with pre-training distributions.
This is evident if we remove entities directly relevant to the answer, such as ``J. Edgar'', and replace them with recursive entity-seeking problems where models must first identify them, shown as ``3 Layers'' in \Cref{fig:sample_question}. We observe that ChatGPT no longer produces the right answer without enough semantic hints and hallucinates ``Adam Sandler'', which may be because ``Eden'' is related ``Adam'', and at the same time, both the writer and main character of ``Gardener of Eden'' is named ``Adam.'' In addition, ChatGPT almost always comes up with ``Adam Sandler'' if asked for actors named ``Adam.''

To investigate and quantify whether LLMs can follow a correct reasoning path and resist the attempt to take ``greedy shortcuts'' during inference, we introduce \modelname~(\fullname), a sophisticated multi-hop question answering dataset, meticulously designed for diminishing semantic associations and gauging the capability of LLMs in undertaking extensively chained reasoning processes. 
To create \modelname, we design a method that removes semantic shortcuts that will lead to the correct answer while adding ``deceptive semantic cues'' that are distracting and irrelevant to the correct answer. We also guarantee a simple and consistent reasoning path and only use common facts that should be memorized by the models. In this way, if an LLM fails on such tasks, it directly suggests that the LLM is taking incorrect shortcuts during inference. For instance, we modify the original question in \Cref{fig:sample_question} by substituting ``J. Edgar'' with a placeholder dubbed ``Film A'' and supplementing a descriptive sentence about the film. Our intent with this strategy is to lessen the model's reliance on direct information from the film name.
We can automatically extend the depth of a reasoning chain by further replacing entities in the descriptive sentence about ``Film A'' with corresponding abstract names. In particular, this generation method is developed on a knowledge base, ensuring that the information employed predates the training data of the examined LLMs.

We then evaluate model performances on the \modelname{} benchmark. We show that state-of-the-art LLMs,\footnote{In this context, ChatGPT and GPT-4.}
are incapable of proper reasoning for instances in \modelname{}, with GPT-4 only achieving 62\% to 64\% accuracy in identifying common entities in Wikipedia and ChatGPT less than 40\%. On the contrary, humans achieve near-perfect performances without much effort because of the simplicity and consistency of the gold reasoning path. We also show that GPT-4 performance strongly correlates with the semantic similarities between the gold answer and other entities mentioned in the question. These findings combined suggest that even the best language models, with detailed in-context-learning (ICL) processes, still fall for deceptive semantic shortcuts and hence hallucinate and fail on \modelname{}.

To summarize, the contribution of our work is three-fold. First, we propose a novel method for generating question-answering data with extended reasoning chains, which allows us to create deceptive semantic shortcuts. Second, we propose \modelname, a QA dataset specifically designed for evaluating LLMs in scenarios with reduced or deceptive semantic associations. 
Finally, our extensive experimental findings suggest that contemporary LLMs predominantly depend on semantic associations in question answering. As entities in question are replaced with alternate inquiries, LLMs cannot adhere to an accurate reasoning process necessary for problem resolution.

\begin{figure*}[t!]
\begin{center}
    \includegraphics[scale=0.34]{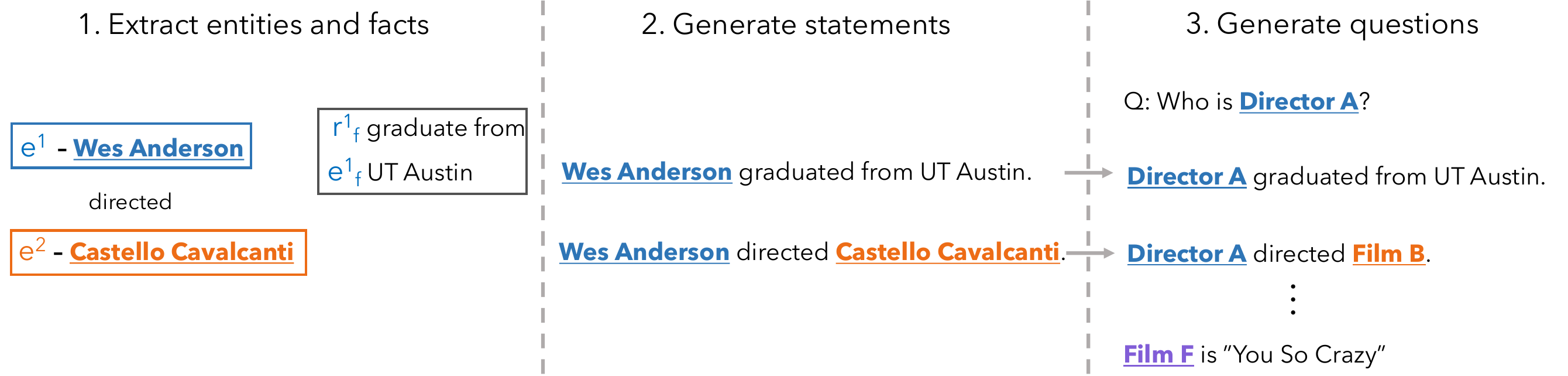}
    \caption{The data generation process of \modelname.}
    \label{fig:generation_process}
    \vspace{-1em}
\end{center}
\end{figure*}

%% file: content/3_dataset.tex
\section{\modelname}

In this section, we introduce the dataset \modelname for evaluating LLMs on extended reasoning chains. We start with the definition and generation approach of a reasoning chain (\Cref{ssec:rChainGen}), followed by the processes for 
question generation (\Cref{ssec:question}) 
and refinement (\Cref{ssec:processing}).

\subsection{Reasoning Chain}\label{ssec:rChainGen}
In \modelname, every question is constructed through an implicit reasoning chain, as depicted in \Cref{fig:generation_process}~(\texttt{1}). This chain is structured in layers, each of which comprises three components: an \textbf{entity} $e_i$, an associated \textbf{fact} $f_i$ about $e_i$, and a relation $r_i$ which connects $e_i$ and the entity in the succeeding layer of the chain, referred to as $e_{i+1}$. The identity of $e_i$ is constrained by $f_i$, $e_{i+1}$ and $r_i$ together. In most cases, neither $f_i$ nor $e_{i+1}/r_i$ can lead to $e_i$, but we guarantee the uniqueness of $e_i$ when subjecting to both constraints. This design allows us to extend the reasoning chain by replacing $e_{i}$ with $f_{i}$, $r_i$, and $e_{i+1}$, until we provide the actual entity name of $e_N$. A valid reasoning path would be to take the provided $e_N$ and resolve $e_{N-1}$ based on $f_{N-1}$ and $r_{N-1}$ until we identify $e_1$, which is the answer to the overall question.

To automatically collect such data, we use a structured knowledge base DBpedia \cite{dbpedia}, because it provides accurate facts and relations between entities, and at the same time, all these facts and relations should have been memorized by most LLMs since Wikipedia is seen during pre-training. 
We use the \emph{2021-09} DBPedia snapshot to guarantee that the information of entities was accessible to LLMs during their training phase. 
The criteria for choosing valid $e_i$, $r_i$, and $f_i$ to form chains are the following:\footnote{Being ``valid'' here refers to satisfying the following criteria for data extraction.}

\stitle{Entity.} Each entity $e_i$ must exist in the knowledge base as an ``entity.''

\stitle{Relation.} Each relation $r_i$ should connect $e_i$ with more than one potential candidate entities $e_{i+1}$. This ensures no direct semantic shortcut between $e_i$ and $e_{i+1}$. For example, the relation ``\texttt{award}'' is deemed valid for the entity ``Tiger Woods'' because he has won multiple awards.
In contrast, the relation ``\texttt{college}'' is invalid because DBPedia only has one corresponding entity, which is \emph{Stanford University}.

\stitle{Fact.} Facts are generated based on DBpedia relations and destination entities for each $e_i$. Specifically, a valid fact $f_i$ consists of a relation $r_i^f$ and a destination entity $e_i^f$. Note that $r_i^f$ is different from $r_i$, because it is only used to generate a single factual statement regarding $e_i$, and not being used to extend the reasoning chain.
We adopt two criteria for fact selection, namely \texttt{easy} and \texttt{hard}. 
The \texttt{easy} criterion selects ($r_i^f$, $e_i^f$) pairs where $e_i$ is the only entity in the database that corresponds to the triplet  (\emph{?entity}, $r_i^f$, $e_i^f$). 
For example, \emph{Jason Connery} is the only entity that fits into (\emph{?entity}, \texttt{father}, \emph{Sean Connery}) because \emph{Jason Connery} is the only child of \emph{Sean Connery}. In this way, $f_i$ itself is sufficient for identifying $e_i$. Contrarily, the \texttt{hard} selection ensures that not only $e_i$ satisfies (\emph{?entity}, $r_i^f$, $e_i^f$). For instance, (\texttt{parentCompany}, \emph{Comcast}) is a \texttt{hard} fact for ``\emph{NBCUniversal}'' since ``\emph{Comcast}'' is also the parent company of ``\emph{Xfinity}.'' The \texttt{hard} criterion ensures that $f_i$ cannot uniquely identify $e_i$. We apply \texttt{easy} and \texttt{hard} criteria to fact generation respectively for \modelname, resulting in two distinct sets of data: \modelname\textsuperscript{easy} and \modelname\textsuperscript{hard}.

\stitle{Chain Construction.}
Following the previously established criteria, we employ a random-walk algorithm to fabricate chains of reasoning. 
The procedure initiates with a designated seed entity $e_1$ in the knowledge base. 
Subsequently, a relation $r_1$, an associated fact $f_1$, and a subsequent entity $e_2$ are chosen randomly to construct a valid layer of the reasoning chain. 
This operation is continuously executed until no more valid layer is found or the cumulative number of layers equals $N_{max}$. We set $N_{max}=5$ in this work. 
We perform at most 50 random walks starting from each seed entity to optimize efficiency and remove duplicated chains. 
Seed entities are sourced via two methods, either through prompting ChatGPT or by data extraction from \href{https://today.yougov.com/ratings/entertainment/fame/people/all}{https://today.yougov.com/}.

This approach ensures that reasoning chain generation for \modelname satisfies the following three properties of each chain:

\stitle{Reasoning-dependent.} Every intermediate layer relies on information from the subsequent layer in the chain for resolution. Consequently, the model must commence from the terminal layer and engage in sequential reasoning throughout the chain.

\stitle{Length-flexible.} The extent of the reasoning chain can be conveniently adjusted by adding or removing layers, thereby enabling an evaluation of the depth of reasoning.

\stitle{Determinism-adjustable.} Determinism of the reasoning chain can be altered by omitting fact $f_i$ in a layer, facilitating an evaluation of the model in addressing questions with multiple potential answers, which is not considered in this work.

\subsection{Question Generation}\label{ssec:question}
In \modelname, each question, denoted as $q$, is a natural language articulation of the above reasoning chains. A layer-by-layer procedure translates this structured chain into a human-readable text. Here we denote $q = q_0 + q_1 + ... + q_n + m_{n+1}$ where $\{q_i|1<=i<=n\}$ indicates the sub-question for the $i$-th layer and ``$+$'' stands for concatenation. $q_0$ and $m_{n+1}$ will be introduced later in this subsection. \Cref{fig:generation_process}~(\texttt{2}) provides a tangible example of this translation process.
Every layer comprises an entity $e_i$, a relation $r_i$, a succeeding entity $e_{i+1}$, and an associated fact $f_i$, which includes the corresponding $r_i^f$ and $e_i^f$. 
The initial step involves translating the triplets, ($e_i$, $r_i$ $e_{i+1}$) and ($e_i$, $r_i^f$, $e_i^f$) into two statements, $s_i^{relation}$ and $s_i^{fact}$, respectively. This is done by few-shot prompting ChatGPT, which we detail in \Cref{ssec:prompt}.

The subsequent step involves substituting each entity $e$ in every statement with a placeholder, as illustrated in  \Cref{fig:generation_process}~(\texttt{3}). In particular, each layer now associates with a pair of statements, namely $s_i^{relation}$ and $s_i^{fact}$, which derive from the translation of the relational triplet ($e_i$, $r_i$ $e_{i+1}$) and the factual triplet ($e_i$, $r_i^f$, $e_i^f$) correspondingly.  In this context, we proceed to solely obfuscate $e_i$ and $e_{i+1}$ in $s_i^{relation}$ and $s_i^{fact}$ by substituting them with their respective hypernyms, $h_i$ and $h_{i+1}$, whilst preserving the identity of $e_i^f$. Hypernym details are extracted from DBpedia if available and sought from the LLM otherwise. ChatGPT is employed in our approach with two few-shot examples, listed in \Cref{ssec:prompt}, obtained from DBpedia, to ensure a consistent level of granularity of hypernym. Each hypernym is subsequently appended with a specific label that is assigned in alphabetical order.
For example, the third layer of a reasoning chain could hold ``Actor C stars in TV Series D'' as $s_3^{relation}$ while $s_3^{fact}$ could be ``Actor C is the artist for \textit{Along on Christmas Day}.'' Here, $e_3$ is substituted by ``Actor C'' and $e_4$, which also appears in the fourth layer, is replaced by ``TV Series D.'' The resulting statements with masked entities are denoted as $m_i^{relation}$ and $m_i^{fact}$ respectively.

The final step involves generating an interrogative question, represented as $q_0$, and an entity information statement, denoted as $m_{n+1}$, pertaining to $q_n$. The selection of the initial interrogative question is contingent upon the hypernym of the entity $e_1$ in the first layer. Specifically, $q_0$ starts with ``Who is ...'' if the hypernym $h_1$ corresponds to a person or ``What is ...'' in all other cases. In the example shown in \Cref{fig:generation_process}, ``Wes Anderson'' has the hypernym ``director.''  The generation process for $m_{n+1}$ is essentially uncomplicated. Given that $m_i^{relation}$ in the terminal layer exhibits the relation between $e_n$ and $e_{n+1}$, and no additional layers exist to identify $e_{n+1}$, we generate $m_{n+1}$ utilizing the template ``$h_{n+1}$ is $e_{n+1}$.'', which is exemplified as ``Film F is \textit{You So Crazy}.'' in \Cref{fig:generation_process}.

\subsection{Question Refinement}\label{ssec:processing}
We perform \textbf{viability filtering} to check if a reasoning chain can be correctly followed to derive the expected answer. As mentioned above, each layer of a specific chain specifies $e_i$ by both a relational triplet ($e_i$, $r_i$ $e_{i+1}$) and a factual triplet ($e_i$, $r_i^f$, $e_i^f$). By excluding $e_i$ from these triplets, we query our knowledge base with both ($?variable$, $r_i$ $e_{i+1}$) and ($?variable$, $r_i^f$, $e_i^f$). The layer is deemed viable if $e_i$ is a unique solution of $?variable$. We only retain those reasoning chains where each layer has passed the viability filtering.

This work aims to analyze models' reasoning capabilities. As such, we employ a \textbf{knowledge filtering} procedure to ensure that most LLMs have sufficient world knowledge to answer our questions. To illustrate this, we consider a question, $q = q_0 + q_1 + ... + q_n + m_{n+1}$, and verify both $s_i^{relation}$ and $s_i^{relation}$, where $1 <= i <= n$. In more explicable terms, we evaluate knowledge by presenting ChatGPT with the prompt ``\emph{Is this statement correct: [$s_i$] Yes or No?}'' A statement is considered as memorized if the LLM response includes a ``Yes.''\footnote{we applied self-consistency strategy \cite{wang2022self} with three runs for a majority vote.} A $q_i$ is deemed to have satisfied the examination criteria if and only if \textit{both} of $s_i^{relation}$ and $s_i^{relation}$ are categorized as memorized.

\begin{figure}[t!]
\begin{center}
    \includegraphics[scale=0.28]{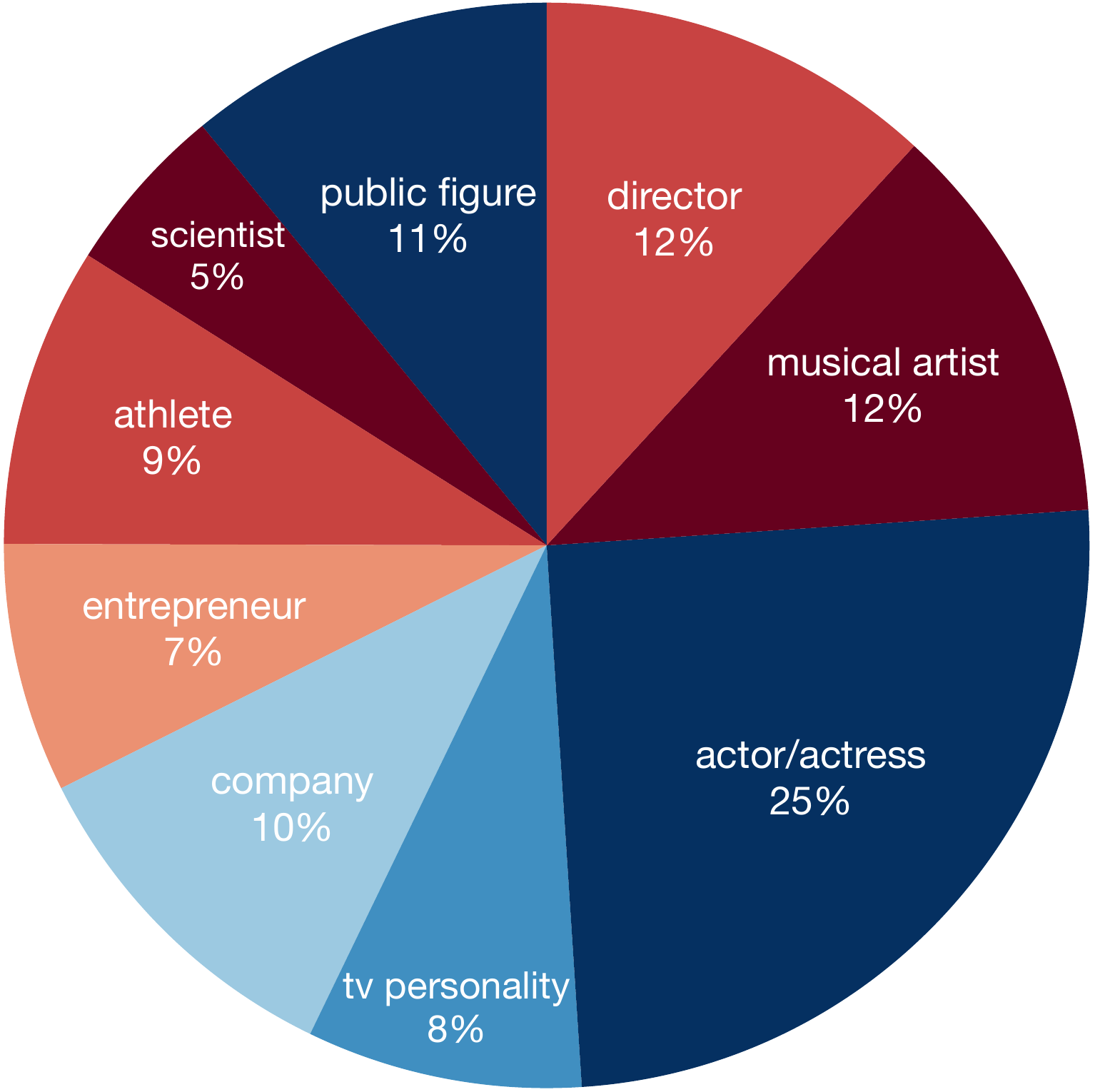}
    \caption{Categorical distribution of seed entities in questions of \modelname.}
    \label{fig:topicStat}
\end{center}
\end{figure}

\input{content/tables/dataStat}

\subsection{Data Statistics}
As indicated in \Cref{table:dataStat}, \modelname encompasses a total of 2,991 questions. The dataset is split into two levels of difficulty: \texttt{hard} and \texttt{easy} according to the criteria discussed in \Cref{ssec:question}. Specifically, the \texttt{easy} set comprises 428 five-layer questions while the \texttt{hard} set includes a larger set of 1,363 five-layer questions. This \texttt{easy} version enables manifesting LLM behavior towards questions with sufficient semantic shortcuts. From this \texttt{hard} set, we randomly select 300 five-layered questions. By removing layers sequentially, the \texttt{hard} set also yields 300 questions in each number of layers from four to one respectively. These \texttt{hard} questions with varying layers allow for a comprehensive assessment of LLMs in terms of their reasoning capabilities across various depths of reasoning. 
The distribution of categories of seed entities for the questions in \modelname is shown in \Cref{fig:topicStat}, showcasing a broad spectrum of topics encompassed by these entities, effectively reducing any potential bias arising from specific entity categories.


%% file: content/tables/dataStat.tex
\begin{table}[t]
\centering
{\small
\begin{tabular}{m{32pt}|m{14pt}|m{15pt}|m{15pt}|m{15pt}|m{15pt}|m{15pt}}
\toprule
Difficulty    & \texttt{easy}    & \texttt{hard}    & \texttt{hard}    & \texttt{hard}    & \texttt{hard}   & \texttt{hard}    \\
\midrule
\#Layers    & 5    & 5    & 4    & 3    & 2    & 1    \\
\midrule
Count    & 428    & 1363    & 300    & 300    & 300    & 300    \\
\bottomrule
\end{tabular}
}
\caption{Statistics of \modelname.}
\label{table:dataStat}
\end{table}

%% file: content/4_expSetup.tex
\input{content/tables/mainRes}
\section{Experiment Setup}
This section presents the experiment configurations employed for assessing the long-chain reasoning capabilities of LLMs through \modelname. This section begins with a discussion on model configurations (\Cref{ssec:modelSet}), subsequently introducing the adopted prompting methodologies (\Cref{ssec:promptMethod}), and finally, it scrutinizes the evaluation configurations (\Cref{ssec:evalSetting}).

\subsection{Model Configuration}\label{ssec:modelSet}
We consider state-of-the-art LLMs to evaluate on \modelname: ChatGPT (gpt-3.5-turbo-0301), Gemini-Pro (Gemini 1.0 Pro) and GPT-4 (gpt-4-0314). All models run with a temperature $\tau$ of 0.8. 

\subsection{Prompting Methods}\label{ssec:promptMethod}
We employ two prompting methods in our experiment: namely, \textit{direct} and \textit{icl}. The \textit{direct} technique presents raw questions to the model without any supplementary context. On the other hand, \textit{icl} method utilizes a few-shot approach, preceding each query with two context-specific examples. Each example is characterized by a sample problem, succeeded by the statement ``Let's solve this question step by step'' and a step-by-step solution written by humans. In these solutions, the problem is analyzed from the final layer through the initial layer to give the correct answer, thereby obviating any necessity for backtracking during the problem-solving procedure. An example is detailed in \Cref{ssec:prompt}.

\subsection{Evaluation Protocol}\label{ssec:evalSetting}
We evaluate the \textbf{accuracy} of LLMs towards the correct answer. Considering that \modelname functions as a free-form QA benchmark and our observation indicates that LLMs typically respond with comprehensive reasoning processes, following prior studies \cite{Agrawal2015VQAVQ,ossowski-hu-2023-retrieving}, we employ a string-match criterion: If the correct answer, identifiable as an entity, is present in the LLM response, we deem such a response as correct. We use a self-consistency of five runs with ChatGPT and Gemini-Pro, but we do not apply self-consistency to GPT-4 due to cost issues.

%% file: content/tables/mainRes.tex
\begin{table*}[t]
\centering
\small
\label{table:methods}
\begin{tabular}{@{}cccccccccccccc@{}} 
\toprule
\multicolumn{11}{c}{\texttt{hard}} & \multicolumn{2}{c}{\texttt{easy}}\\
\cmidrule(l){2-11}\cmidrule(l){12-13}
depth &\multicolumn{2}{c}{\texttt{d}=1} & \multicolumn{2}{c}{\texttt{d}=2} & \multicolumn{2}{c}{\texttt{d}=3} & \multicolumn{2}{c}{\texttt{d}=4} & \multicolumn{2}{c}{\texttt{d}=5} & \multicolumn{2}{c}{\texttt{d}=5} \\ 
\cmidrule(r){2-3} \cmidrule(lr){4-5} \cmidrule(l){6-7} \cmidrule(l){8-9} \cmidrule(l){10-11}\cmidrule(l){12-13} 
& \textit{direct} & \textit{icl} & \textit{direct} & \textit{icl} & \textit{direct} & \textit{icl} & \textit{direct} & \textit{icl} & \textit{direct} & \textit{icl}& \textit{direct} & \textit{icl} \\ 
\midrule
ChatGPT & 22.3 & 53.3 & 7.0   & 40.0   & 5.0   & 39.2   & 3.7   & 39.3   & 7.2   & 39.0     &13.1 &47.0  \\

Gemini-Pro & 45.0 & 49.3 & 29.5   & 23.5   & 27.3   & 28.6   & 25.7   & 24.3   & 17.2   & 21.5     &30.6 &38.9  \\

GPT-4 & 60.3   & 76.0    & 50.0   & 63.7   & 51.3   & 61.7   & 52.7   & 63.7  & 46.9   & 61.9  & 66.4&81.8\\
\bottomrule
\end{tabular}
\caption{Accuracy of ChatGPT, Gemini-Pro and GPT-4 across different depths \texttt{d} of reasoning (number of layers in the questions) as well as the difficulty of the questions. We evaluate two prompt strategies: \textit{direct} zero-shot prompt and \textit{icl} with two examples. }

\label{table:mainRes}
\end{table*}


%% file: content/5_results.tex
\section{Results}

The results of the state-of-the-art LLMs on \modelname, as reported in \Cref{table:mainRes}, illustrate the variations in performance across different depths of reasoning and levels of difficulty. In general, with the entities recursively substituted by the descriptions of reasoning chaining layers, and therefore eliminating surface-level semantic cues, these models generate more incorrect answers. When the reasoning depth increases from one to five on \texttt{hard} questions, there is a notable decline in performance for all models, with an average accuracy decrease of 14.2\% when using the \textit{icl} prompt and 14.3\% when using the \textit{direct} prompt. The performance of LLMs is significantly higher on the \texttt{easy} set compared to the \texttt{hard} set. This is evidenced by a marked increase in accuracy, averaging 7.0\% for ChatGPT, 15.4\% for Gemini-Pro, and 19.7\% for GPT-4. This finding underscores the significant impact that semantic shortcuts have on the accuracy of responses, and it also indicates that GPT-4 is considerably more capable of identifying and taking advantage of these shortcuts.

It can also be inferred from the results that the demonstrated human-written examples do help improve the performance. For both ChatGPT and GPT-4, using the \textit{icl} prompt consistently leads to better performance than the \textit{direct} prompt. This suggests that these models can benefit from examples that provide explicit and well-crafted human-written reasoning processes. These few-show examples encourage the models to resolve the questions with the correct reason instead of guessing with semantic cues. Nonetheless, even with clear reasoning processes provided, models still tend to fail for incorrect shortcuts.

Another observation is that LLMs do not always perform worse on questions with more depth of reasoning. For instance, ChatGPT with \textit{direct} prompt is 3.5\% more accurate when the depth of reasoning \texttt{d} increases from four to five. We believe this also indicates that LLMs do not always follow the reasoning path and may instead answer the question based on surface-level semantic shortcuts. Further analyses can be found at \Cref{ssec:entSim}

To identify the errors in these LLM responses, we randomly sampled 20 data points where GPT-4 with \textit{icl}, the best performing method in this benchmark, has generated incorrect answers. The most frequent error pattern lies in hallucination during the reasoning process, encompassing 80\% of the cases. In such cases, an intermediate reasoning stage erroneously conceived an inaccurate, albeit plausible answer, culminating in accumulated errors in the final answer. For instance, the ``founder of a rocket company'' was misconstrued as ``\emph{Elon Musk}, the founder of SpaceX'', rather than the correct answer of ``\emph{Jeff Bezos}, the founder of Blue Origin.'' This phenomenon proves our claim that LLM relies on semantic cues for problem-solving instead of cognitive reasoning. 

\section{Analysis and Discussions}
In this section, we perform further analyses 
and provide a more in-depth discussion of our findings. 

\subsection{Can human solve \modelname?}
Towards a more comprehensive insight into human proficiency on \modelname, a human analysis was carried out focusing on the \texttt{hard} set characterized by a reasoning depth of five. This evaluation involves two computer science PhD students as the annotators to ensure their expertise. A selection of 50 questions, randomly extracted from the set, was subjected to annotation. Each question received annotations from both annotators independently, ensuring a dual perspective on every query. During the annotation process, annotators are presented with an input question to which they are required to provide an answer. To facilitate this, they are granted access to conduct information searches through 
the \emph{dbpedia-snapshot-2021-09} database. This could be achieved either through visiting the specific entity page or by querying via a SPARQL portal. The averaged accuracy of annotators achieves 95\% with an Inter-annotator Agreement of Cohen $\kappa=0.79$.

\subsection{Do LLMs take Shortcuts?} 
\label{ssec:entSim}

\begin{figure}[t!]
    \includegraphics[scale=0.52]{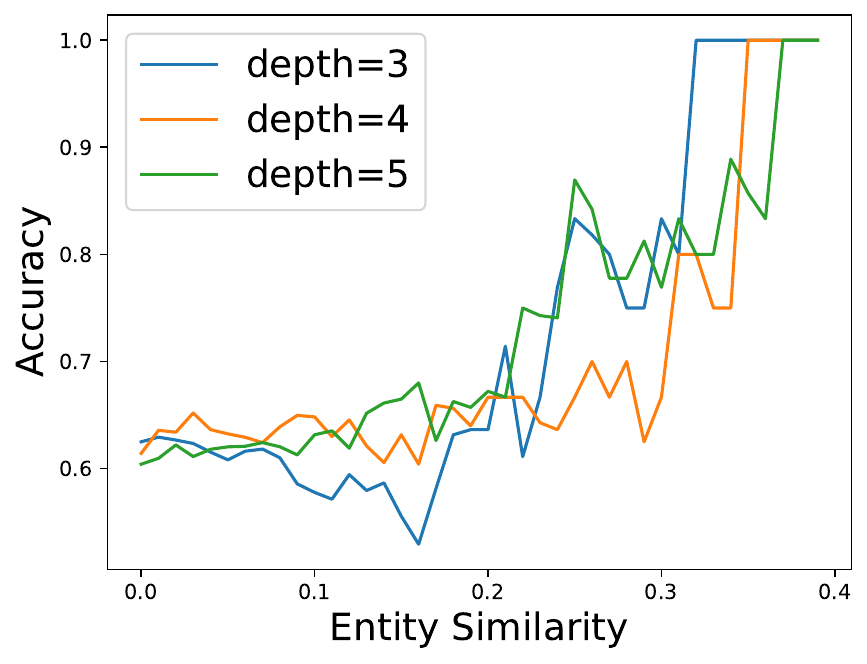}
    \caption{The correlation between GPT-4 performance on \modelname hard set and entity similarities.}
    \label{fig:similarity-performance}
\end{figure}

\begin{figure}[t!]
    \includegraphics[scale=0.52]{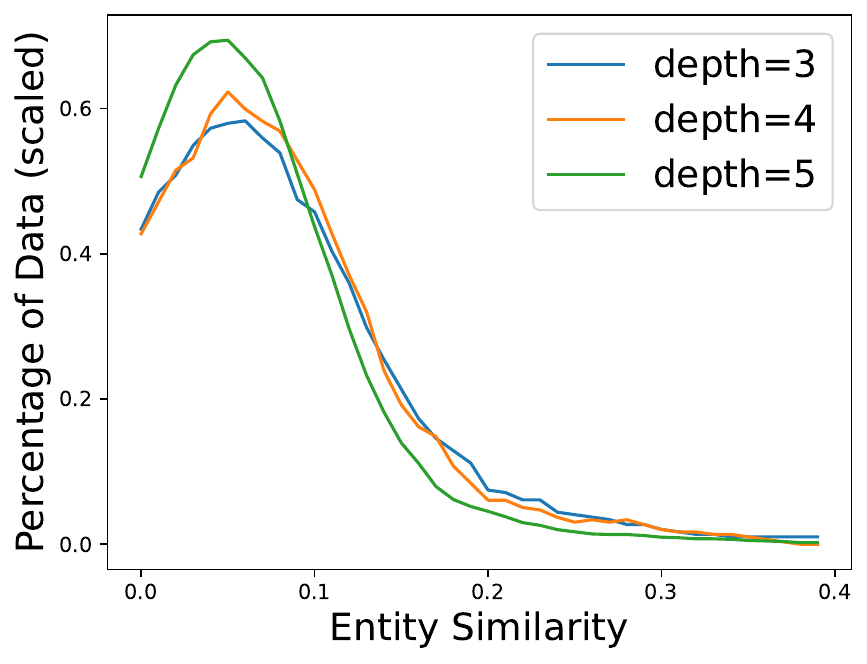}
    \caption{The distribution of entity similarity scores.}
    \label{fig:similarity-distribution}
\end{figure}

\begin{figure}[t!]
    \includegraphics[scale=0.52]{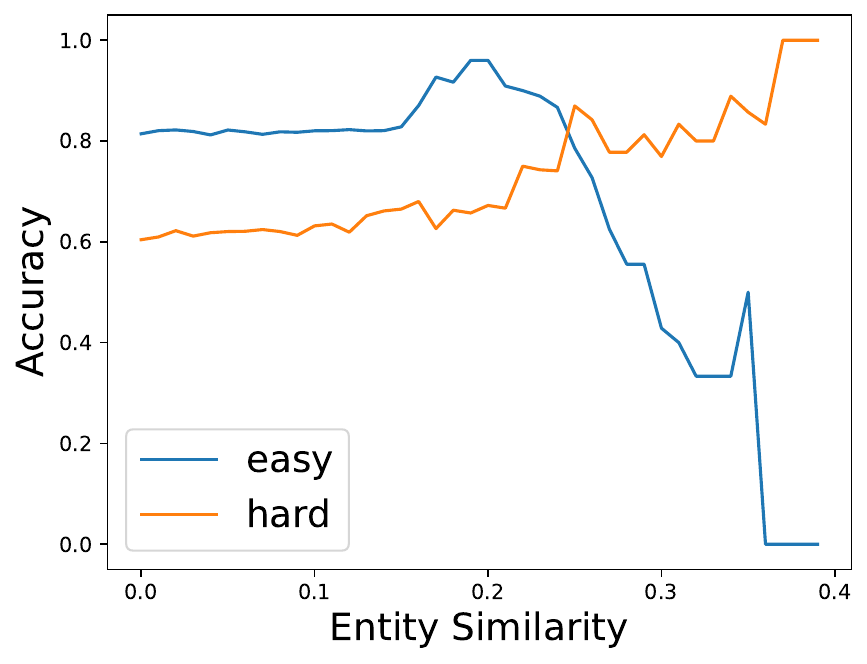}
    \caption{GPT-4 performances with different entity similarity scores between the easy and hard sets.}
    \label{fig:similarity-easy-hard}
\end{figure}

One of our main motivations is to test if language models can follow a simple yet effective reasoning chain instead of taking semantic shortcuts based on entity associations. We design an analytical experiment based on entity similarity to investigate whether LLMs take such semantic shortcuts. Our intuition is to model the correlation between the performances and the averaged semantic similarities between the gold answer and other entities mentioned in the question. In an ideal situation where a model takes an optimal reasoning path for each instance, we will see a uniform distribution of performances: the accuracy for various degrees of entity similarities will be relatively the same. If the model relies much on entity biases and takes semantic shortcuts, we will see an increasing performance when the mentioned entities are more closely related to the target answer. 

To infer entity similarities, we employ an off-the-shelf sentence Transformer model\footnote{\url{https://huggingface.co/sentence-transformers/msmarco-distilbert-cos-v5}.} and encode the entity strings into embeddings. With such embeddings, we calculate the dot-product similarity between the target answer and all other Wikipedia entities mentioned in the question and compute an averaged similarity for each instance. We then draw the correlation curve of model performances on instances with certain similarities.\footnote{Specifically, we start with a similarity value $x_{start}=0.0$ and increment it by a step of $0.01$. For each $x_{start}$, we find instances that have an averaged similarity score between $[x_{start}, x_{start}+0.1]$, and compute the accuracy on these instances.} This method derives a relatively continuous curve based on our limited evaluation data.

\stitle{Observation 1.} Fig.~\ref{fig:similarity-performance} shows that GPT-4's performance positively correlates with the entity similarities in the instances. This shows that the model relies on semantic shortcuts instead of following the correct reasoning path. 

\stitle{Observation 2.} GPT-4 performs similarly on different depths if the entity similarity scores are the same, especially on instances with lower similarities ($<0.25$). This again suggests that the model relies on spurious entity biases to solve the question, because it does not do better on instances with shorter reasoning paths. That being said, the performance differences between lower and higher depths mostly come from the different data distributions, as shown in Fig.~\ref{fig:similarity-distribution}.

\stitle{Observation 3.} Fig.~\ref{fig:similarity-easy-hard} shows another interesting finding, where GPT-4 does better on easy instances with lower similarity scores ($<0.25$) and performs worse on those with higher similarity scores. Our explanation is that GPT-4 tends to take more obvious shortcuts by relying on only one or two entities in the easy instance, and ``early exits'' without even considering the entities that are closely relevant to the gold answer.

\subsection{Do open source LLMs perform better?}\label{ssec:llamaRes}
To expand the scope of our findings, we experimented with open-sourced LLAMA-2 models across different sizes on hard questions of \modelname and their results can be found at \Cref{table:llamaRes}. Similar to our observations on GPT-series models, there’s a notable decline in the accuracy of Llama models as the reasoning depth increases from one to five on the hard set. We’ve also conducted the entity similarity analysis which led to the same observation as Observation 1 in \Cref{ssec:entSim}: LLMs’ performance positively correlates with the entity similarities in the instances. These conclusions strengthened our claim that models rely on semantic shortcuts instead of following the correct reasoning path.
\subsection{Will prompting solve \modelname{}?}
Although the effectiveness of prompting techniques is out of the scope of this paper, we still additionally tested the Tree of Thought(TOT)\cite{yao2023tree} method on ChatGPT with a ``propose'' strategy, which tries to decompose the questions layer by layer and solve them sequentially. Our results show that such a prompting method fails completely even if we provide human-written examples for question decomposition. In no experiment did the TOT method generate a valid answer in the final response. 

That being said, it is unlikely that there exists a prompting technique that will significantly improve models' performance on \modelname over CoT, as long as we do not enforce the correct reasoning chain (which defeats the purpose of testing generalizability) during inference.

\subsection{Will optimal retrieval solve \modelname?}\label{ssec:optRAG}
Although our knowledge filtering process has already removed the knowledge barrier, it can still be pointed out that whether Retrieval Augmented Generation(RAG) method can solve this task. To address such concerns, we tested GPT-4 on 300 randomly sampled 5-layer hard questions. To minimize the impact of the retrieval method, we investigate a ``performance upper bound'' setting, which directly injects the retrieval result of the visible entities and relations in the input question from DBpedia. To be specific, statements like \texttt{``Storm Warning (1951 film) stars Actor D''} will have \texttt{``<Storm Warning (1951 film), starring, ?>: Ginger Rogers, Steve Cochran, Ronald Reagan, Doris Day''} prepended to the input question. The retrieved knowledge will obtain at most 20 candidate entities and the correct entity is guaranteed to be included. A complete example of a question and retrieved knowledge can be found in \Cref{fig:fig2}. We believe this represents an upper bound of the retrieval result. Additionally, we provide each input with a 1-shot demonstration with the retrieved knowledge, input questions, and human-written reasoning thoughts. 

The accuracy of GPT-4 through the above process is 62.0\%, which is close to our reported performance of 61.9\% in the original setting. We can then hypothesize that simply injecting knowledge into the model can hardly solve the problem and the bottleneck remains at the reasoning/question decomposition ability of LLMs. Moreover, it can be concluded that our knowledge-filtering process can effectively estimate parametric knowledge, and it is also reliable after self-consistency.

%% file: content/2_relatedWork.tex
\section{Related Work}

We discuss two topics of works that are highly relevant to this study.

\stitle{Hallucination.} Existing research on analyzing LLMs hallucination mainly focuses on three aspects: input-conflicting hallucination, context-conflicting hallucination, and fact-conflicting hallucination \cite{zhang2023siren}. 
Input-conflicting hallucination happens when the LLMs outputs are divergent from prompts \cite{maynez2020faithfulness}. 
Self-contradictory outputs can occur in long-form or multi-turn answers, implying that LLMs lose track of the core input information during generation \cite{liu2023lost}. Factual-based errors -- mis-aligns with established world knowledge -- are most widely studied since they can happen in various LLMs to affect multiple tasks' performance \cite{min2023factscore,li2023halueval}.
Knowledge conflicts also lead to hallucination when in-context information contradicts what LLMs memorize from pre-raining \cite{longpre-etal-2021-entity,zhou-etal-2023-context,wang-etal-2023-causal}.
However, none of these existing works look into hallucination on reasoning-required questions and test whether the models are following the correct reasoning path versus the semantic token bias.

\stitle{Reasoning capability of LLMs.} The success of in-context learning \cite{brown2020language} and instruction tuning \cite{wei2021finetuned} have inspired various works to solve reasoning tasks by prompting LLMs. 
More advanced prompting strategies have been proposed to enhance the reasoning capabilities of LLMs. These investigations take advantage of the heuristic nature of human problem-solving procedures and incorporate them into textual prompts as guidance for the models. 
An example of this can be seen in chain-of-thought prompting \cite{wei2022chain}, a method in which intermediate steps are generated leading up to a final solution. \citet{yao2023tree} and \citet{zhou2022least} proposed to decompose the questions into simpler, manageable sub-problems as a method to facilitate complex reasoning. 
Researchers also attempted to investigate LLMs' reasoning ability over structured knowledge \cite{ding2023knowledge} which necessitates reasoning with uncertainty as well as back-tracking. Studies prior to LLMs also propose to generate rationales to improve with more faithful reasoning \cite{rajani-etal-2019-explain,wang2022pinto}.
Nonetheless, this work questions the exhibited ``reasoning capabilities'' of LLMs by distinguishing them from the attempt to take greedy semantic shortcuts during the reasoning process.

%% file: content/6_conclusion.tex
\section{Conclusion}
This paper proposes a novel QA benchmark for probing LLM hallucinations induced by semantic shortcuts on reasoning chains. We introduced a systematic method for generating question-answering data with extended reasoning chains. This dataset enables us to examine the reasoning capabilities of LLMs on extended reasoning paths and our experimental results indicate that LLMs predominantly depend on semantic shortcuts for reasoning and such behavior contributes to its failure. Our analyses questioned the validity of current LLMs and also inspired future studies. The problem-solving or reasoning trajectories of humans could be incorporated into the training or inference phase of current Language Modeling systems for more effective computational models.

\section*{Ethical Considerations}
Innovations in technology often encounter the moral challenge of dual-use: the same development can bring both benefits and risks. With the probing method and benchmark presented in this paper, the line between beneficial and harmful usage largely depends on data. Proper utilization of the technology necessitates the legal and ethical acquisition of input text corpora and other modalities of inputs. Legal frameworks and standards are crucial for ensuring proper data use and for granting individuals the right to remove their data. In the absence of such regulation, the ethical use of data depends on the responsibility of technology users. Additionally, the generated and analysis data may exhibit biases that systematically affect accuracy for less represented groups or in new areas, potentially resulting in performance disparities among sub-populations based on ethnicity, race, gender, and other factors. Moreover, the effectiveness of trained systems diminishes when applied to new data that deviates from their training set. Therefore, issues of generalizability and fairness must be thoroughly examined when implementing the methodologies discussed in this paper. It is crucial to embed ethical considerations as fundamental principles at each stage of system development, ensure high levels of transparency and clarity in data, algorithms, models, and functions within the system, release software under open-source licenses to facilitate public scrutiny and investigate strategies to safeguard at-risk groups.

\section*{Limitations}
Our work proposes an analytical framework and dataset to evaluate how well language models can take the right reasoning paths instead of deceptive semantic shortcuts. To this end, we identify the following limitations. 

\stitle{Limited Baselines.} We only consider two variants of language models as our baselines. With more efforts in the future, we can extend to more families of large language models with different prompting techniques. Although we project that all current systems will be far behind human performances on \modelname{}, we may identify specific models and methods that are more resistant to deceptive semantic shortcuts.

\stitle{Limited Entities.} We only consider popular entities in Wikipedia as our target answer. Future works may benefit from considering a wider range of entities and a more general setting of our data extraction processes. 

\section*{Acknowledgement}

We appreciate the reviewers for their insightful
comments and suggestions.
Bangzheng Li is supported by the Faculty Startup Fund of UC Davis, and the Provost's Fellowship.
Fei Wang is supported by the Annenberg Fellowship and the Amazon ML Fellowship.
Ben Zhou was funded by ONR Contract N00014-23-1-2365.
Xingyu Fu was funded by NSF grant IIS-2212433.
Muhao Chen is supported by the NSF Grant IIS 2105329, the NSF Grant ITE 2333736, the Faculty Startup Fund of UC Davis, a Cisco Research Award and two Amazon Research Awards.

\noindent\paragraph{}

%% file: content/7_appendix.tex
\clearpage
\section{Appendices}
\begin{figure}[h]
\begin{center}
    \includegraphics[scale=0.28]{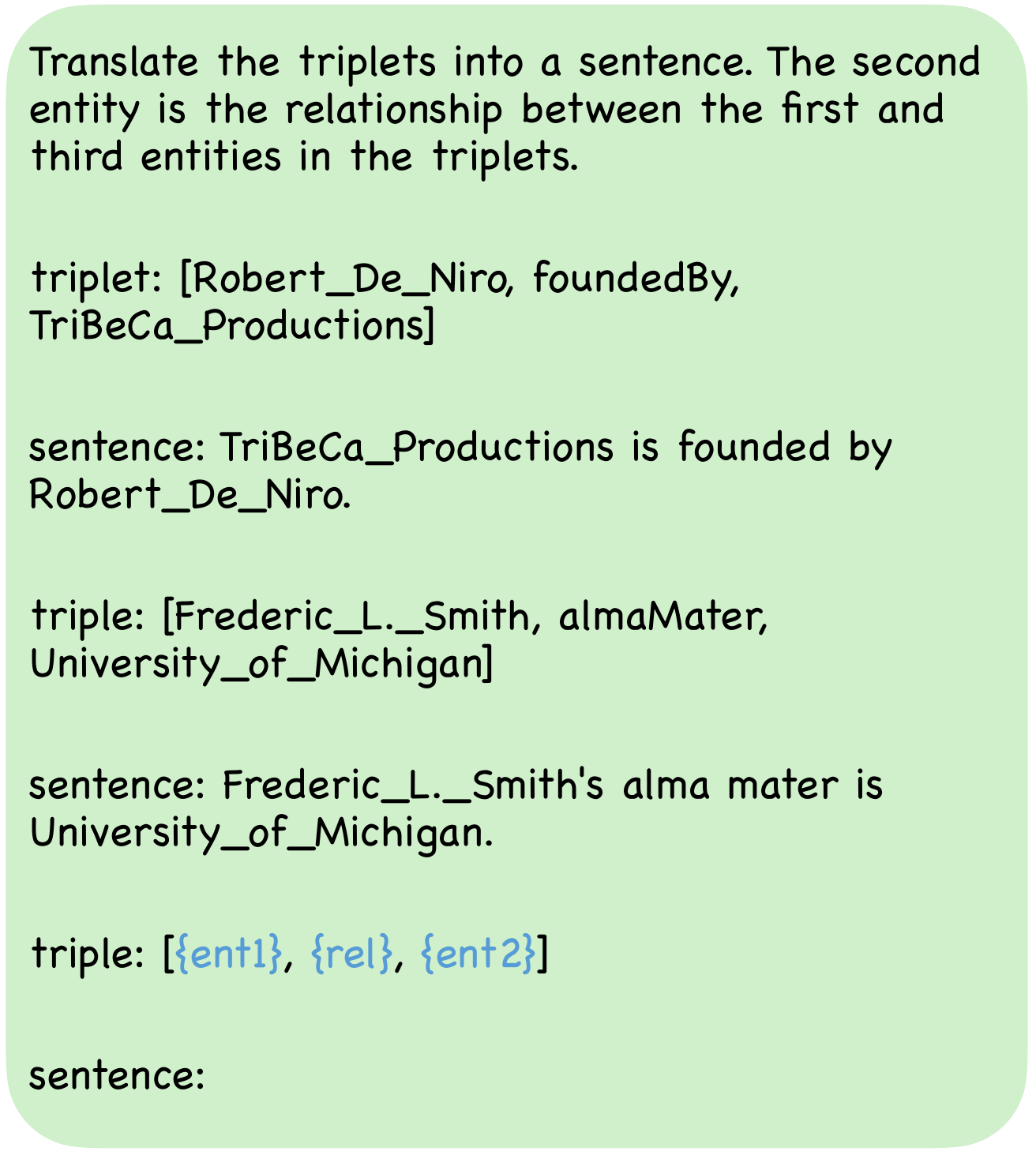}
    \caption{Verbalization Prompt: verbalize (\emph{entity}, \texttt{relation}, \emph{entity}) into a natural sentence.}
    \label{fig:tripletPrompt}
\end{center}
\end{figure}

\begin{figure}[h]
\begin{center}
    \includegraphics[scale=0.28]{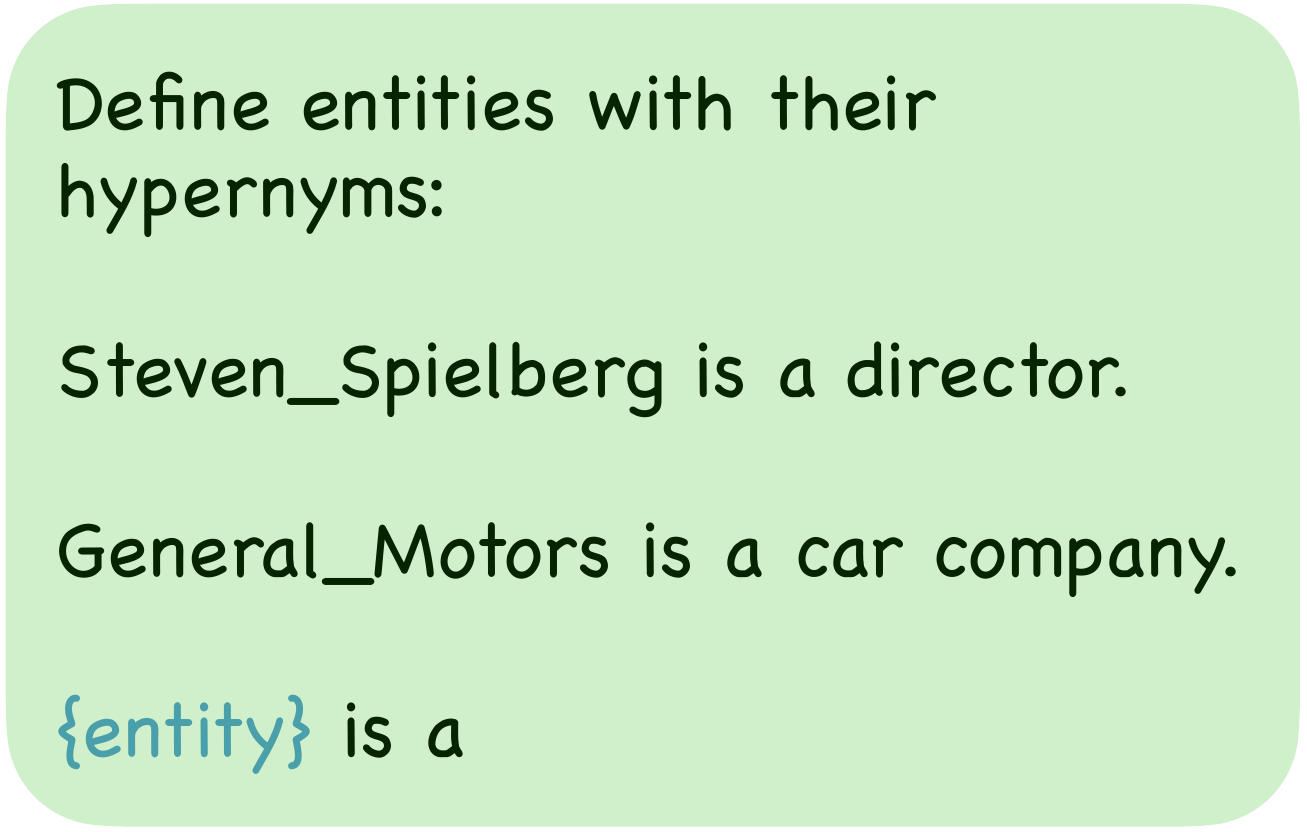}
    \caption{Typing Prompt: generate hypernym of an entity when it is unavailable from DBpedia.}
    \label{fig:typingPrompt}
\end{center}
\end{figure}

\begin{figure}[h]
\begin{center}
    \includegraphics[scale=0.28]{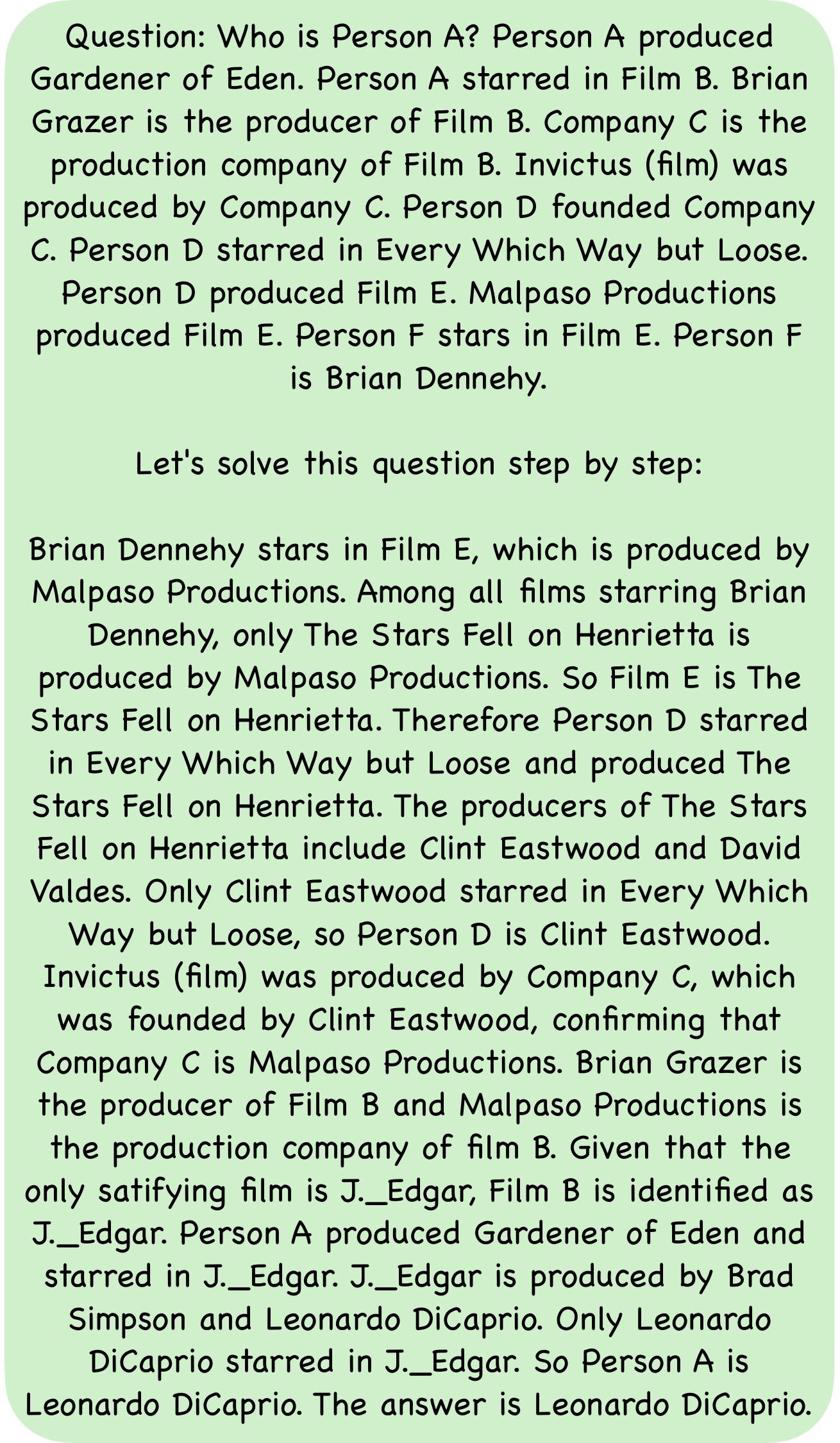}
    \caption{One of the two examples used in \textit{icl} prompt, with depth of five reasoning process.}
    \label{fig:ansPrompt}
\end{center}
\end{figure}

\begin{figure}[h]
\begin{center}
    \includegraphics[scale=0.28]{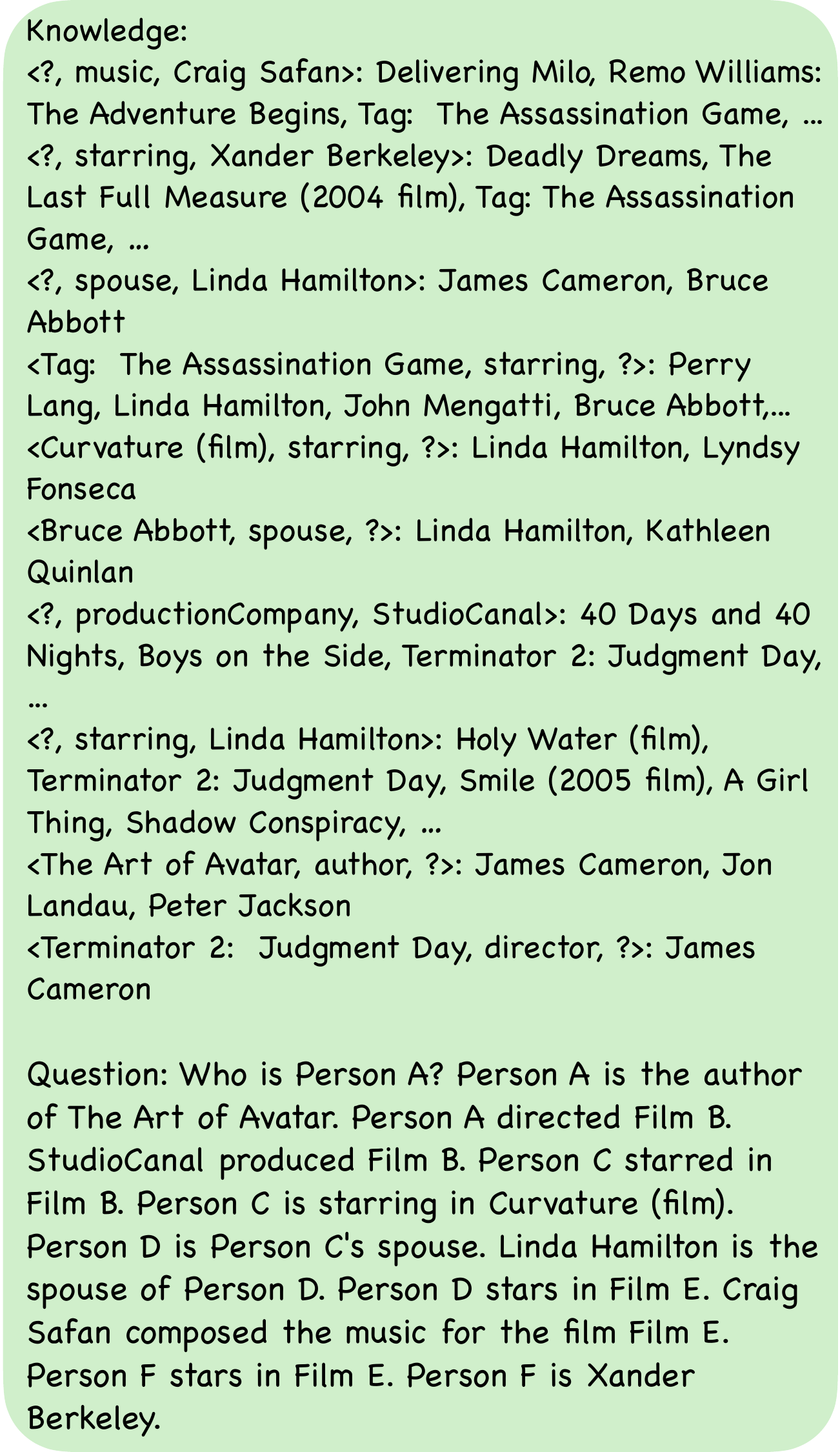}
    \caption{Sample knowledge and question in the ``performance uppper bound'' setting of RAG study}
    \label{fig:ragExp}
\end{center}
\end{figure}

\appendix
\label{ssec:prompt}
We present the prompts used in data generation or model evaluation on \modelname. During the question generation process (\Cref{ssec:question}), The Verbalization Prompt in \Cref{fig:tripletPrompt} transforms a triplet, ($e_i$, $r_i$ $e_{i+1}$) or ($e_i$, $r_i^f$, $e_i^f$), into a statement by querying ChatGPT with it. Typing Prompt in \Cref{fig:typingPrompt} is later used when we try to substitute an entity with its hypernyms in a statement but its hypernym is unavailable from DBpedia. \Cref{fig:ansPrompt} is one of the two examples we used in the \textit{icl} prompt answering questions with a depth of five. Questions with fewer depths of reasoning adopt two similar examples with corresponding depths. \Cref{fig:ragExp} is an example input of the RAG experiment, as elaborated in \Cref{ssec:optRAG} and illustrated in \Cref{fig:fig2}.

\Cref{table:llamaRes} is the experimental results of the open-sourced Llama-2 models on \modelname{}, as discussed in \Cref{ssec:llamaRes}.

\begin{table*}[h]
\centering
{
\begin{tabular}{@{}cccccccccccccc@{}} 
\toprule
depth &\multicolumn{2}{c}{\texttt{d}=1} & \multicolumn{2}{c}{\texttt{d}=2} & \multicolumn{2}{c}{\texttt{d}=3} & \multicolumn{2}{c}{\texttt{d}=4} & \multicolumn{2}{c}{\texttt{d}=5} \\ 
\cmidrule(r){2-3} \cmidrule(lr){4-5} \cmidrule(l){6-7} \cmidrule(l){8-9} \cmidrule(l){10-11} 
& \textit{direct} & \textit{icl} & \textit{direct} & \textit{icl} & \textit{direct} & \textit{icl} & \textit{direct} & \textit{icl} & \textit{direct} & \textit{icl}\\ 
\midrule
Llama-2-7b & 29.5   & 22.0   & 21.4   & 11.9   & 15.8   & 13.0   & 10.6   & 11.3   & 10.9   & 14.5   \\
Llama-2-13b & 40.0   & 28.5  & 27.0   & 16.4   & 20.5   & 15.8   & 11.3   & 15.9   & 13.9   & 19.4   \\
Llama-2-70b & 45.0   & 57.0   & 24.5   & 35.2   & 17.8   & 39.0   & 18.5   & 32.5   & 12.1   & 34.5   \\
\bottomrule
\end{tabular}
}
\caption{Accuracy of Llama-2 models across different depths \texttt{d} of reasoning (number of layers in the questions). We evaluate two prompt strategies: \textit{direct} zero-shot prompt and \textit{icl} with two examples. The 7b and 13b versions are chat variants and the 70b version is instruct variant.}

\label{table:llamaRes}
\end{table*}